\documentclass[11pt,a4paper]{article}
\usepackage[hyperref]{acl2020}
\usepackage{times}
\usepackage{latexsym}

\usepackage{microtype}
\usepackage{amsmath,amsfonts,amssymb,bbm,epsfig,bm}
\usepackage{algorithm}
\usepackage{algorithmicx}
\usepackage{algpseudocode}
\usepackage{amsmath}
\usepackage{xr-hyper}
\usepackage{multirow}
\usepackage{graphicx}  
\aclfinalcopy 
\def\aclpaperid{838} 

\title{Reasoning Over Semantic-Level Graph for Fact Checking}
\author{Wanjun Zhong$^1$\thanks{\ \ \ Work done while this author was an intern at Microsoft Research.} , Jingjing Xu$^3$$^*$, Duyu Tang$^2$, Zenan Xu$^1$, Nan Duan$^2$, Ming Zhou$^2$\\
	\bf Jiahai Wang$^1$ and Jian Yin$^1$\\
	$^1$ The School of Data and Computer Science, Sun Yat-sen University.\\
	Guangdong Key Laboratory of Big Data Analysis and Processing, Guangzhou, P.R.China\\
	$^2$ Microsoft Research $^3$ MOE Key Lab of Computational Linguistics, School of EECS, Peking University\\
	{\tt \{zhongwj25@mail2,xuzn@mail2,wangjiah@mail,issjyin@mail\}.sysu.edu.cn}\\
	{\tt \{dutang,nanduan,mingzhou\}@microsoft.com}\\ 
	{\tt jingjingxu@pku.edu.cn}\\
}

\date{}

\begin{document}
\maketitle
\begin{abstract}
Fact checking is a challenging task because verifying the truthfulness of a claim requires reasoning about multiple retrievable {evidence}. In this work, we present a method suitable for reasoning about the semantic-level structure of evidence. Unlike most previous {works}, which typically {represent} evidence sentences with either string concatenation or fusing the features of isolated evidence sentences, our approach operates on rich semantic structures of evidence obtained by semantic role labeling. We propose two mechanisms to exploit the structure of evidence while leveraging the advances of pre-trained models like BERT, \mbox{GPT} or XLNet. Specifically, using XLNet as the backbone, we first utilize the graph structure to re-define the relative distances of words, with the intuition that semantically related words should have short distances. Then, we adopt graph convolutional network and graph attention network to propagate and aggregate information from neighboring nodes on the graph. We evaluate our system on FEVER, a benchmark dataset for fact checking, and find that rich structural information is helpful and both our graph-based mechanisms improve the accuracy. Our model is the state-of-the-art system in terms of both official evaluation metrics, namely claim verification accuracy and FEVER score.
	\end{abstract}

\section{Introduction}
Internet provides an efficient way for individuals and organizations to quickly spread information to massive audiences. 
However, 
malicious people spread false news, which may have significant influence on public opinions, stock prices, even presidential elections \cite{faris2017partisanship}.
\newcite{vosoughi2018spread} show that false news reaches more people than the truth.
The situation is more urgent as advanced pre-trained language models \cite{radford2019language} can produce remarkably coherent and fluent texts, which lowers the barrier for the abuse of creating deceptive content.
In this paper, we study fact checking
with the goal of automatically 
assessing the truthfulness
of a textual claim by looking for textual evidence.


\begin{figure}[t]
	\centering
	\includegraphics[width=0.49\textwidth]{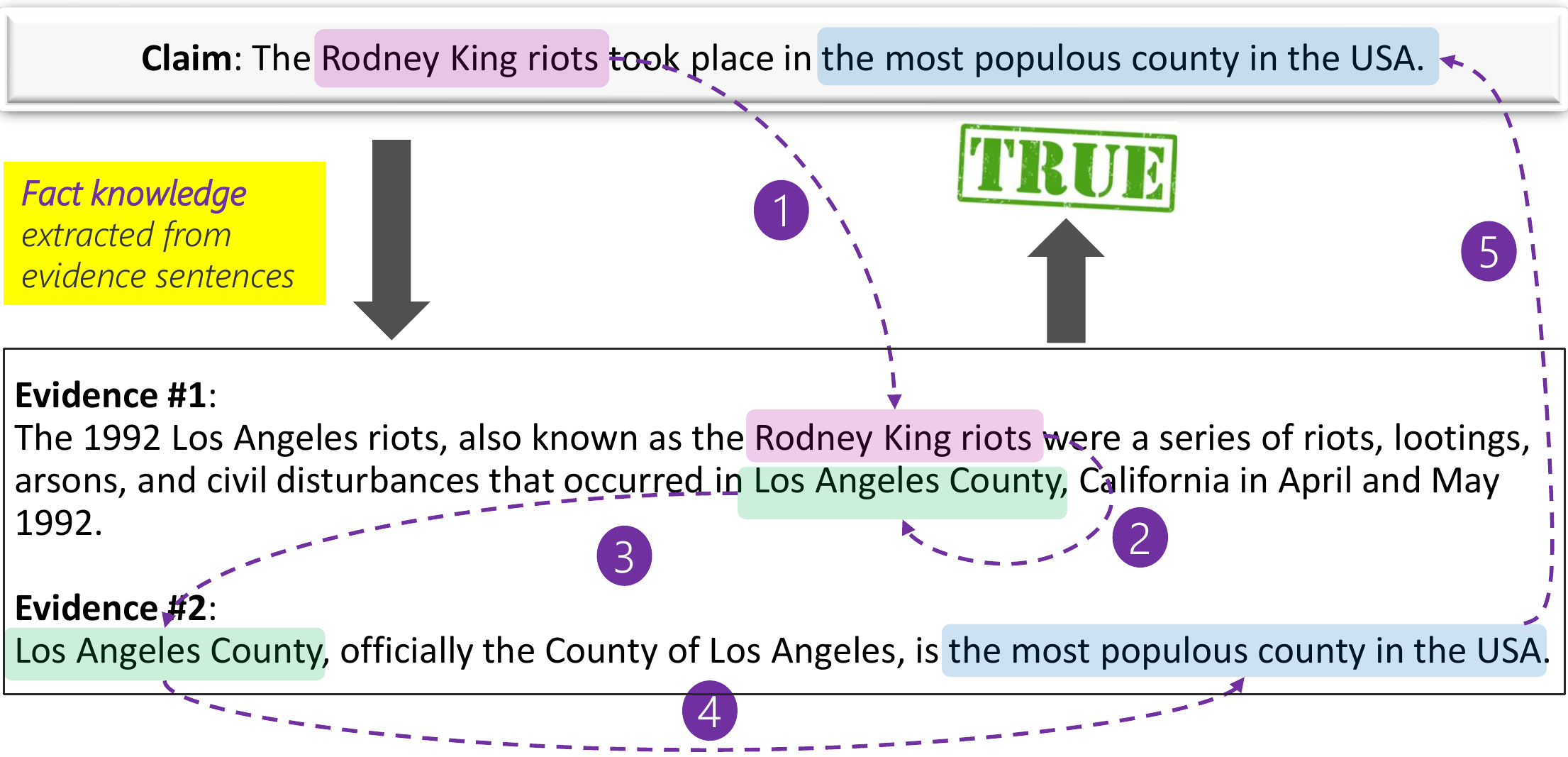}
	\caption{A motivating example for fact checking and the FEVER task. Verifying the claim requires understanding the semantic structure of multiple evidence sentences and the reasoning process over the structure.}
	\label{fig:task-example}
\end{figure}

Previous works are dominated by natural language inference models \cite{dagan2013recognizing,angeli2014naturalli} because the task requires reasoning of the claim and retrieved evidence sentences.
They typically either concatenate evidence sentences into a single string, which is used in top systems in the FEVER challenge \cite{thorne2018fact}, or use  feature fusion to aggregate the features of isolated evidence sentences
\cite{zhou-etal-2019-gear}.
However, both methods fail to capture rich semantic-level structures among multiple evidence, which also prevents the use of deeper reasoning model for fact checking.
In Figure \ref{fig:task-example}, we give a motivating example.
Making the correct prediction requires a model to reason based on the understanding that ``\textit{Rodney King riots}'' is occurred in ``\textit{Los Angeles County}'' from the first evidence, and that ``\textit{Los Angeles County}'' is ``\textit{the most populous county in the USA}'' from the second evidence.
It is therefore desirable to mine the semantic structure of evidence and leverage it to verify the truthfulness of the claim.

Under the aforementioned consideration, we present a graph-based reasoning approach for fact checking.
With a given claim, we represent the retrieved evidence sentences as a graph, and then use the graph structure to guide the reasoning process.
Specifically, 
we apply 
semantic role labeling (SRL)
to parse each evidence sentence, and establish links between arguments to construct the graph. 
%
When developing the reasoning approach, we intend to simultaneously leverage rich semantic structures of evidence embodied in the graph and powerful contextual semantics learnt in pre-trained model like BERT \cite{devlin2018bert}, GPT \cite{radford2019language} and XLNet \cite{yang2019xlnet}.
To achieve this, we first  
re-define the distance between words based on the graph structure when producing contextual representations of words.
Furthermore, we 
adopt graph convolutional network and graph attention network to propagate and aggregate information over the graph structure. In this way, the reasoning process employs semantic representations at both word/sub-word level and graph level.

We conduct experiments on FEVER \cite{thorne2018fever}, 
which is one of the most influential benchmark datasets for fact checking.
FEVER consists of 185,445 verified claims, and evidence sentences for each claim are natural language sentences from Wikipedia. 
We follow the official evaluation protocol of FEVER, and demonstrate that
our approach achieves state-of-the-art performance in terms of both claim classification accuracy and FEVER score.
Ablation study shows that the integration of graph-driven representation learning mechanisms improves the performance.
We briefly summarize our contributions as follows.
\begin{itemize}
	\item We propose a graph-based reasoning approach
	for fact checking. Our system apply SRL to construct graphs and present two graph-driven representation learning mechanisms.
	\item Results verify that both graph-based mechanisms improve the accuracy, and our final system achieves state-of-the-art performance on the FEVER dataset.
\end{itemize}


\section{Task Definition and Pipeline}
\label{section:task-and-pipeline}
With a textual claim given as the input, the problem of fact checking is to find supporting evidence sentences to verify the truthfulness of the claim.

We conduct our research on FEVER \cite{thorne2018fever}, short for Fact Extraction and VERification, a benchmark dataset for fact checking.
Systems are required to retrieve evidence sentences from Wikipedia, and predict the  
claim as ``\textsl{SUPPORTED}", ``\textsl{REFUTED}" or ``\textsl{NOT ENOUGH INFO (NEI)}", standing for that the claim is supported by the evidence, refuted by the evidence, and is not verifiable, respectively.
There are two official evaluation metrics in FEVER.
The first is the accuracy for three-way classification.
The second is FEVER score, which further measures the percentage of correct retrieved evidence for ``\textsl{SUPPORTED}'' and ``\textsl{REFUTED}'' categories.
Both the statistic of FEVER dataset and the equation for calculating FEVER score are given in Appendix B.  

\begin{figure}[h]
	\includegraphics[width=0.5\textwidth]{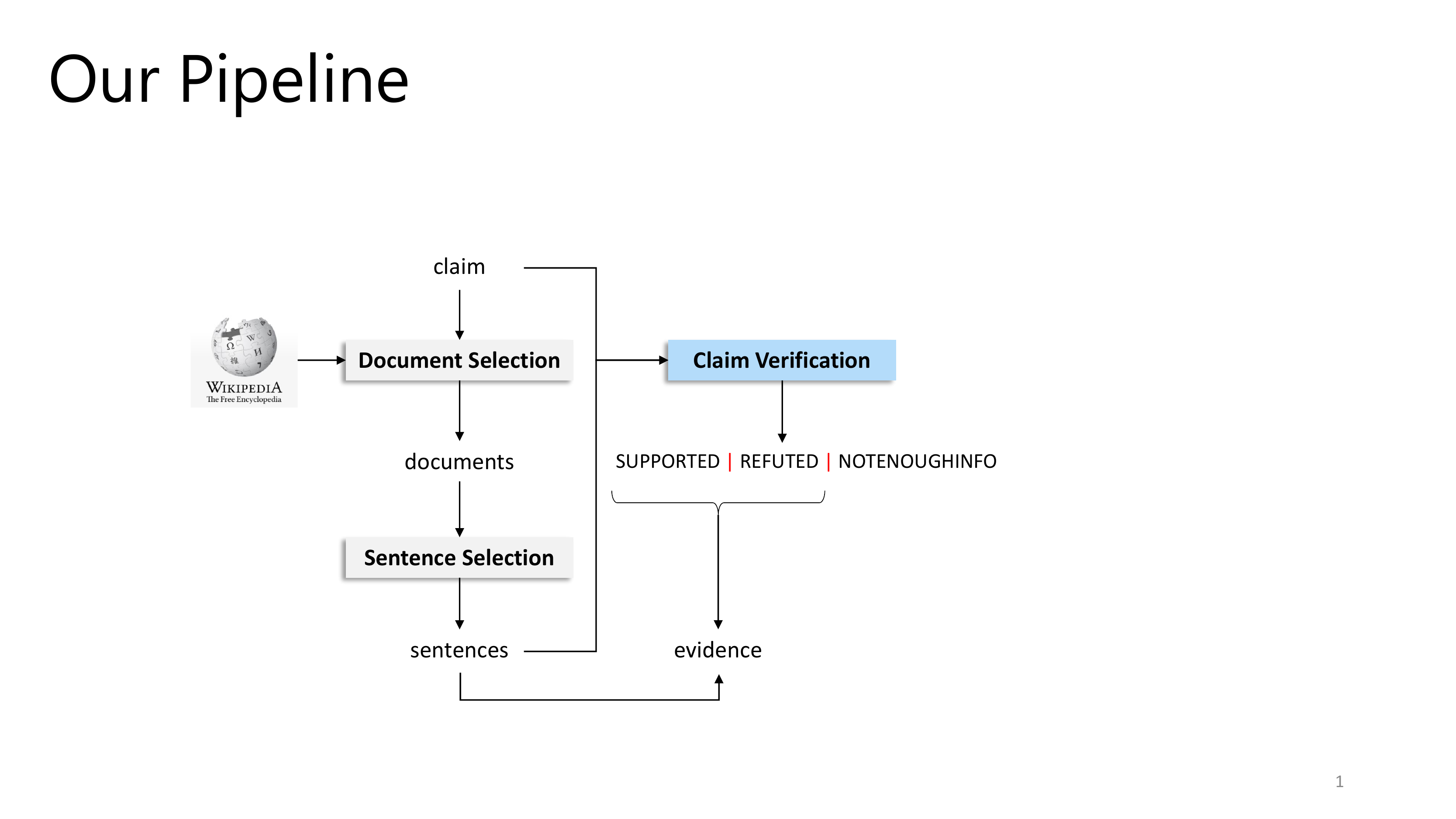}
	\caption{Our pipeline for fact checking on FEVER. The main contribution of this work is a graph-based reasoning model for claim verification. }
	\label{fig:overall-pipeline}
\end{figure}
\begin{figure*}[t]
	\centering
	\includegraphics[width=\textwidth]{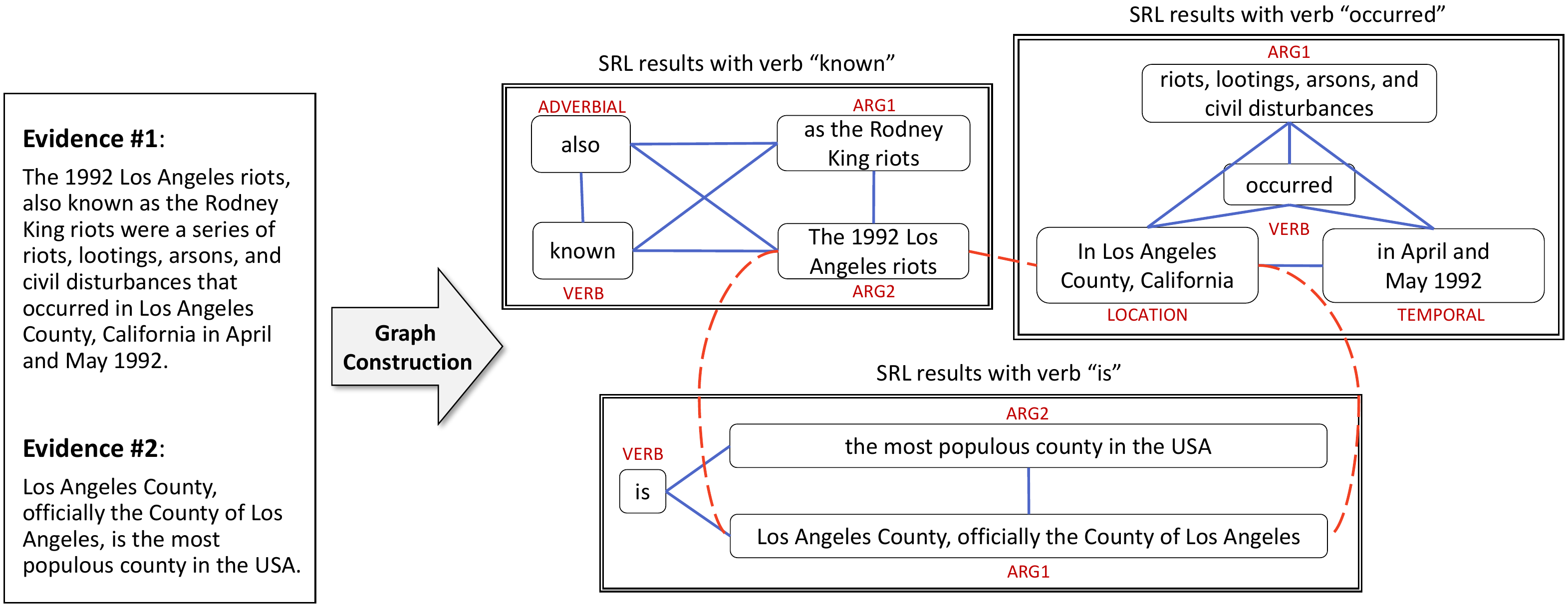}
	\caption{The constructed graph for the motivating example with two evidence sentences. Each box describes a ``tuple'' which is extracted by SRL triggered by a verb. Blue solid lines indicate edges that connect arguments within a tuple and red dotted lines indicate edges that connect argument across different tuples.}
	\label{fig:graph-construction}
\end{figure*} 
Here, we present an overview of our pipeline for FEVER, which follows the majority of previous studies.
Our pipeline consists of three main components: a {document retrieval model}, a {sentence-level evidence selection model}, and a {claim verification model}.  
Figure \ref{fig:overall-pipeline} gives an overview of the pipeline.
 With a given claim, the {document retrieval model} retrieves the most related documents from a given collection of Wikipedia documents. 
With retrieved documents, the {evidence selection model} selects top-$k$ related sentences as the evidence. 
Finally, the {claim verification model} takes the claim and evidence sentences and outputs the 
 veracity of the claim. 

The main contribution of this work is the graph-based reasoning approach for claim verification, which is explained detailedly in Section \ref{section:claim-verification}.
Our strategies for document selection and evidence selection are described in Section \ref{sec:document-and-sentence-retrieval}.

\section{Graph-Based Reasoning Approach}
\label{section:claim-verification}
In this section, we introduce our graph-based reasoning approach for claim verification, which is the main contribution of this paper. 
Taking a claim and retrieved evidence sentences\footnote{Details about how to retrieve evidence for a claim are described in Section \ref{sec:document-and-sentence-retrieval}.} as the input, our approach predicts the truthfulness of the claim. For FEVER, it is a three-way classification problem, which predicts the claim as ``\textsl{SUPPORTED}", ``\textsl{REFUTED}" or ``\textsl{NOT ENOUGH INFO (NEI)}".

The basic idea of our approach is to employ the intrinsic structure of evidence to assess the truthfulness of the claim.
As shown in the motivating example in Figure \ref{fig:task-example}, making the correct prediction needs good understanding of the semantic-level structure of evidence and the reasoning process based on that structure.
In this section, we first describe our graph construction module (\S \ref{section:graph-construction}).
Then, we present how to apply graph structure for fact checking, including a contextual representation learning mechanism with graph-based distance calculation (\S \ref{sec:graph-based distance}), and  graph convolutional network and graph attention network to propagate and aggregate information over the graph (\S \ref{section:gcn} and \S \ref{sectin:graph-attention-net}). 

\begin{figure*}[t]
	\includegraphics[width=\textwidth]{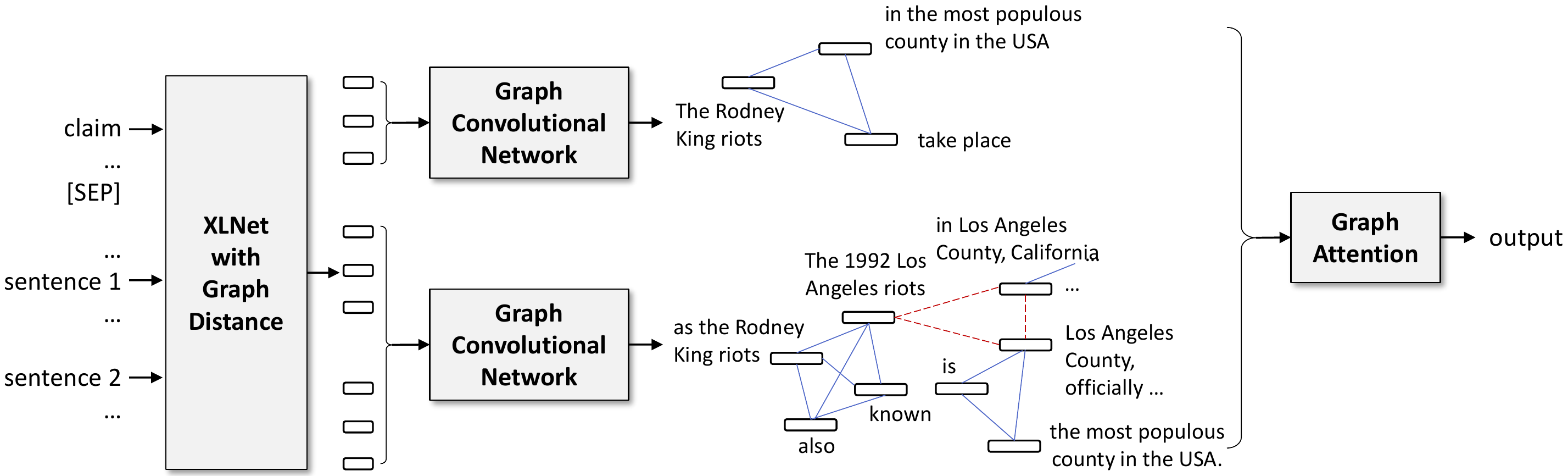}
	\caption{An overview of our graph-based reasoning approach for claim verification. Taking a claim and evidence sentences as the input, we first calculate contextual word representations with graph-based distance (\S \ref{sec:graph-based distance}). After that, we use graph convolutional network to propagate information over the graph (\S \ref{section:gcn}), and use graph attention network to aggregate information (\S \ref{sectin:graph-attention-net}) before making the final prediction.}
	\label{fig:pipeline}
\end{figure*}

\subsection{Graph Construction}
\label{section:graph-construction}
Taking evidence sentences as the input, we would like to build a graph to reveal the intrinsic structure of these evidence.
There might be many different ways to construct the graph, such as open information extraction \cite{banko2007open}, named entity recognition plus relation classification, sequence-to-sequence generation which is trained to produce structured tuples \cite{goodrich2019assessing}, etc. 
In this work, we adopt a practical and flexible way based on semantic role labeling \cite{carreras2004introduction}. 
Specifically, with the given evidence sentences, our graph construction operates in the following steps. 
\begin{itemize}
	\item 
	For each sentence, we parse it to tuples\footnote{A sentence could be parsed as multiple tuples.} with an off-the-shelf SRL toolkit developed by AllenNLP\footnote{\url{https://demo.allennlp.org/semantic-role-labeling}}, which is a re-implementation of a BERT-based model \cite{shi2019simple}.
	\item For each tuple, we regard its elements with certain types as the nodes of the graph.
	We heuristically set those types as verb, argument, location and temporal, which can also be easily extended to include more types.
	We create edges for every two nodes within a tuple.
	\item We create edges for nodes across different tuples to capture the structure information among multiple evidence sentences.
	Our idea is to create edges for nodes that are literally similar with each other.
	Assuming entity $A$ and entity $B$ come from different tuples,  we add one edge if one of the following conditions is satisfied: (1) $A$ equals $B$; (2) $A$ contains $B$; (3) the number of overlapped words between $A$ and $B$ is larger than the half of the minimum number of words in $A$ and $B$.
\end{itemize}
Figure~\ref{fig:graph-construction} shows the constructed graph of the evidence in the motivating example.
In order to obtain the structure information of the claim, we use the same pipeline to represent a claim as a graph as well.

Our graph construction module offers an approach on modeling structure of multiple evidence, which could be further developed in the future.

\subsection{Contextual Word Representations with Graph Distance}
\label{sec:graph-based distance}
We describe the use of graph for learning graph-enhanced contextual representations of words\footnote{In Transformer-based representation learning pipeline, the basic computational unit can also be word-piece. For simplicity, we use the term ``word'' in this paper.}.

Our basic idea is to shorten the distance between two semantically related words on the graph, 
which helps to enhance their relationship 
when we calculate contextual word representations with a Transformer-based \cite{vaswani2017attention} pre-trained model like BERT and XLNet.
Suppose we have five evidence sentences \{$s_1$, $s_2$, ... $s_5$\} and the word $w_{1i}$ from $s_1$ and the word $w_{5j}$ from $s_5$ are connected on the graph,
simply concatenating evidence sentences as a single string fails to capture their semantic-level structure, and would give a large distance to $w_{1i}$ and $w_{5j}$, which is the number of words between them across other three sentences (i.e., $s_2$, $s_3$, and $s_4$).
An intuitive way to achieve our goal is to define an $N \times N$ matrix of distances of words along the graph, where $N$ is the total number of words in the evidence.
However, this is unacceptable in practice because the representation learning procedure will take huge memory space, which is also observed by \newcite{shaw2018self}. 

In this work, we adopt pre-trained model XLNet \cite{yang2019xlnet} as the backbone of our approach because it naturally involves the concept of relative position\footnote{Our approach can also be easily adapted to BERT by adding relative position like \newcite{shaw2018self}.}.
Pre-trained model captures rich contextual representations of words, which is helpful for our task which requires sentence-level reasoning.
Considering the aforementioned issues, we implement an approximate solution to trade off between the efficiency of implementation and the informativeness of the graph.
Specifically,
we reorder evidence sentences with a topology sort algorithm with the intuition that closely linked nodes should exist in neighboring sentences. This would prefer
%
%
that neighboring sentences contain either parent nodes or sibling nodes, so as to better capture the semantic relatedness between different evidence sentences.
We present our implementation in Appendix A. The algorithm begins from nodes without incident relations. 
For each node without incident relations, we recursively visit its child nodes in a depth-first searching way.
After obtaining graph-based relative position of words, we feed
the sorted sequence
into XLNet to obtain the contextual representations. Meanwhile, we obtain the representation $h([CLS])$ for a special token $[CLS]$, which stands for the \mbox{joint} representation of the claim and the evidence in Transformer-based architecture.

\subsection{Graph Convolutional Network}
\label{section:gcn}
We have injected the graph information in Transformer and obtained $h([CLS])$, which captures the semantic interaction between the claim and the evidence at word level
\footnote{By ``word" in ``word-level", we mean the basic computational unit in XLNet, and thus $h([CLS])$ capture the sophisticated interaction between words via multi-layer multi-head attention operations. }.
As shown in our motivating example in Figure \ref{fig:task-example} and the constructed graph in Figure \ref{fig:graph-construction}, the reasoning process needs to operate on span/argument-level, where the basic computational unit typically consists of multiple words like ``\textit{Rodney King riots}'' and ``\textit{the most popular county in the USA}''. 


%

To further exploit graph information beyond word level, we first calculate the representation of a node, which is a word span in the graph, by averaging the contextual representations of words contained in the node.
After that, we employ multi-layer graph convolutional network (GCNs) \cite{kipf2016semi} to update the node representation by aggregating representations from their neighbors on the graph.
%
Formally, we denote $G$ as the graph constructed by the previous graph construction method and make $H \in \bm{R}^{N^v \times d}$ a matrix containing representation of all nodes, where $N^v$ and $d$ denote the number of nodes and the dimension of node representations, respectively. 
Each row $H_i \in \bm{R}^d$ is the representation of node $i$. 
We introduce an adjacency matrix $A$ of graph $G$ and its degree matrix $D$, where we add self-loops to matrix $A$ and $D_{ii}= \sum_{j} A_{ij}$. 
One-layer GCNs will aggregate information through one-hop edges, which is calculated as follows: 
\begin{equation}
H^{(1)}_i=\rho(\widetilde{A}H_iW_0),
\end{equation}
where $H^{(1)}_i \in \bm{R}^d$ is the new $d$-dimension representation of node $i$, $\widetilde{A}=D^{-\frac{1}{2}}AD^{-\frac{1}{2}}$ is the normalized symmetric adjacency matrix, $W_0$ is a weight matrix, and $\rho$ is an activation function. 
To exploit information from the multi-hop neighboring nodes, we stack multiple GCNs layers:
\begin{equation}
H^{(j+1)}_i=\rho(\widetilde{A}H_i^{(j)}W_j),
\end{equation}
where $j$ denotes the layer number and $H^0_i$ is the initial representation of node $i$ initialized from the contextual representation. We simplify $H^{(k)}$ as $\bm{H}$ for later use, where $\bm{H}$ indicates the representation of all nodes updated by $k$-layer GCNs.

The graph learning mechanism will be performed separately for claim-based and evidence-based graph. 
Therefore, we denote $\bm{H}_c$ and $\bm{H}_e$ as the representations of all nodes in claim-based graph and evidence-based graphs, respectively. 
Afterwards, we utilize the graph attention network to align the graph-level node representation learned for two graphs before making the final prediction.

\subsection{Graph Attention Network}
\label{sectin:graph-attention-net}
We explore the related information between two graphs and make semantic alignment for final prediction.
Let $\bm{H}_e \in \bm{R}^{N^v_e \times d}$ and $\bm{H}_c \in \bm{R}^{N^v_c \times d}$ denote matrices containing representations of all nodes in evidence-based and claim-based graph respectively, where $N^v_e$ and $N^v_c$ denote number of nodes in the corresponding graph. 
\par
We first employ a graph attention mechanism \cite{velivckovic2017graph} to generate a claim-specific evidence representation for each node in claim-based graph. Specifically, we first take each $h^i_c \in\bm{H_c}$ as query, and take all node representations $h^j_e \in \bm{H_e}$ as keys. 
We then perform graph attention on the nodes, an attention mechanism 
$a: \bm{R}^d \times \bm{R}^d \rightarrow \bm{R}$ to compute attention coefficient as follows:
\begin{equation}
e_{ij} = a(\bm{W_c}h^i_c,\bm{W_e}h^j_e)
\end{equation} 
which means the importance of evidence node $j$ to the claim node $i$. 
$W_c\in \bm{R}^{F\times d}$ and $W_e\in \bm{R}^{F\times d}$ is the weight matrix and $F$ is the dimension of attention feature. 
We use the dot-product function as $a$ here.
We then normalize $e_{ij}$ using the softmax function:
\begin{equation}
\alpha_{ij} = softmax_j(e_{ij}) = \frac{exp(e_{ij})}{\sum_{k\in N^v_e} exp(e_{ik})}
\end{equation}
After that, we calculate a claim-centric evidence representation $\bm{X} = [x_1,\dots,x_{N^v_c}]$ using the weighted sum over $\bm{H_e}$:
\begin{equation}
x_i = \sum_{j\in N^v_e}\alpha_{ij}h^j_e
\end{equation}
We then perform node-to-node alignment and calculate aligned vectors $A=[a_1,\dots, a_{N_c^v}]$ by the claim node representation $\bm{H^c}$ and the claim-centric evidence representation $\bm{X}$,
\begin{equation}
a_{i} = f_{align}(h_c^i,x^i),
\end{equation}
where $f_{align}()$ denotes the alignment function. Inspired by \newcite{shen2018improved}, we design our alignment function as:
\begin{equation}
f_{align}(x, y) = W_a[x,y,x-y,x\odot y],
\end{equation}
where $W_a \in \bm{R}^{d \times 4*d}$ is a weight matrix and $\odot$ is element-wise Hadamard product. 
The final output $g$ is obtained by the mean pooling over $A$. 
We then feed the concatenated vector of $g$ and the final hidden vector $h([CLS])$ from XLNet through a MLP layer for the final prediction.

\section{Document Retrieval and Evidence Selection}
In this section, we briefly describe our \mbox{document} retrieval and evidence selection components to make the paper self contained.
\label{sec:document-and-sentence-retrieval}
\subsection{Document Retrieval}
\label{sec:document-retrieval}
The document retrieval model takes a claim and a collection of Wikipedia documents as the input, and returns $m$ most relevant documents. 

We mainly follow \newcite{nie2019combining}, 
the top-performing system on the 
FEVER shared task \cite{thorne2018fact}. 
The document retrieval model first uses {keyword matching} to filter candidate documents from the massive Wikipedia documents.
Then, NSMN \cite{nie2019combining} is applied to handle the documents with disambiguation titles, which are 10\% of the whole documents.
Documents without disambiguation title are assigned with higher scores in the resulting list.
The input to the NSMN model includes
the claim and candidate documents with disambiguation title.
At a high level, NSMN model has 
encoding, alignment, matching and output layers. 
Readers who are interested are recommended to refer to the original paper for more details.


Finally, we select top $10$ documents from the resulting list.

\subsection{Sentence-Level Evidence Selection}
Taking a claim and all the sentences from retrieved documents as the input,
evidence selection model returns the top $k$ most relevant sentences.
\par 
We regard evidence selection as a semantic matching problem, and leverage rich contextual representations embodied in pre-trained models like {XLNet} \cite{yang2019xlnet} and RoBERTa \cite{liu2019roberta} to measure the relevance of a claim to every evidence candidate.
Let's take XLNet as an example.
The input of the sentence selector is \[ce_i = [Claim, SEP, Evidence_i, SEP, CLS]\] where $Claim$ and $Evidence_i$ indicate tokenized word-pieces of original claim and $i^{th}$ evidence candidate, $d$ denotes the dimension of hidden vector, and $SEP$ and $CLS$ are symbols indicating ending of a sentence and ending of a whole input, respectively.
The final representation $h_{\bm{ce_i}} \in \bm{R}^d$ is obtained via extracting the hidden vector of the $[CLS]$ token. 

After that, we employ an MLP layer and a softmax layer to compute score $s^{+}_{ce_i}$ for each evidence candidate.
Then, we rank all the evidence sentences by score $s^{+}_{ce_i}$. The model is trained on the training data with a standard cross-entropy loss. Following the official setting in FEVER, we select top $5$ evidence sentences. The performance of our evidence selection model is shown in Appendix \ref{appendix:evidence-selection}.

\section{Experiments}
We evaluate on FEVER \cite{thorne2018fever}, a benchmark dataset for fact extraction and verification. 
Each instance in FEVER dataset consists of a claim, groups of ground-truth evidence from Wikipedia and a label (i.e., 
``\textsl{SUPPORTED}", ``\textsl{REFUTED}" or ``\textsl{NOT ENOUGH INFO (NEI)}"),
indicating its veracity. 
FEVER includes a dump of Wikipedia, which contains 5,416,537 pre-processed documents. 
The two official evaluation metrics of FEVER are {label accuracy} and {FEVER score}, as described in Section \ref{section:task-and-pipeline}.
Label accuracy is the primary evaluation metric we apply for our experiments because it directly measures the performance of the claim verification model. 
We also report FEVER score for comparison, which measures whether both the predicted label and the retrieved evidence are correct. 
No evidence is required if the predicted label is \textsl{NEI}.


\subsection{Baselines}

We compare our system to the following baselines, including three top-performing systems on FEVER shared task, a recent work GEAR \cite{zhou-etal-2019-gear}, and
a concurrent work by \newcite{liu2019kernel}.
%
%
%
%
%

	\begin{itemize}
		\item \newcite{nie2019combining} employ a semantic matching neural network for both evidence selection and claim verification. 
		\item \newcite{yoneda2018ucl} infer the veracity of each claim-evidence pair and make final prediction by aggregating multiple predicted labels.
		\item  \newcite{hanselowski2018ukp} encode each claim-evidence pair separately, and use a pooling function to aggregate features for prediction.
		\item GEAR \cite{zhou-etal-2019-gear} uses BERT to obtain claim-specific representation for each evidence sentence, and applies graph network by regarding each evidence sentence as a node in the graph.
		\item KGAT \cite{liu2019kernel} is concurrent with our work, which regards sentences as the nodes of a graph and uses Kernel Graph Attention Network to aggregate information. 
	\end{itemize}



\subsection{Model Comparison}




Table \ref{tab:overall-performance} reports the performance of our model and baselines on the blind test set with the score showed on the public leaderboard\footnote{The public leaderboard for perpetual evaluation of FEVER is \url{https://competitions.codalab.org/competitions/18814\#results}. DREAM is our user name on the leaderboard.}.
As shown in Table \ref{tab:overall-performance}, in terms of label accuracy, our model significantly outperforms previous systems with 76.85\% on the test set.
It is worth noting that, our approach, which exploits explicit graph-level semantic structure of evidence obtained by SRL, outperforms GEAR and KGAT, both of which regard sentences as the nodes and use model to learn the implicit structure of evidence
\footnote{We don't overclaim the superiority of our system to GEAR and KGAT only comes from the explicit graph structure,  because we have differences in other components like sentence selection and the pre-trained model.}.
By the time our paper is submitted, our system achieves state-of-the-art performance in terms of both evaluation metrics on the leaderboard. 


\begin{table}[t]
	\begin{tabular}{p{3.86cm}p{1.33cm}p{1.63cm}}
		\hline
		\multirow{2}{*}{Method}   & {Label}         & {FEVER}      \\
		& {Acc (\%)} & {Score~(\%)} \\ \hline
		\newcite{hanselowski2018ukp}                    & 65.46                  & 61.58               \\
		\newcite{yoneda2018ucl} & 67.62                  & 62.52               \\
		\newcite{nie2019combining}                    & 68.21                  & 64.21               \\
		GEAR \cite{zhou-etal-2019-gear}              & 71.60                  & 67.10               \\ 
		KGAT \cite{liu2019kernel} & 72.81 & 69.40\\
		DREAM (our approach)              & \textbf{76.85}         & \textbf{70.60}      \\ \hline
	\end{tabular}
	\caption{Performance on the blind test set on FEVER. Our approach is abbreviated as DREAM.}
	\label{tab:overall-performance}
\end{table}
\subsection{Ablation Study}
Table \ref{tab:ablation-study} presents the label accuracy on the development set after eliminating different components (including the graph-based relative distance (\S \ref{sec:graph-based distance}) and graph convolutional network and graph attention network (\S \ref{section:gcn} and \S \ref{sectin:graph-attention-net}) separately in our model.
The last row in Table \ref{tab:ablation-study} corresponds to the baseline where all the evidence sentences are simply concatenated as a single string, where no explicit graph structure is used at all for fact verification.
\begin{table}[h]
	\centering
	\begin{tabular}{l|c}
		\hline
		Model                  & Label Accuracy \\ \hline
		DREAM               & 79.16      \\
		-w/o Relative Distance & 78.35       \\
		-w/o GCN\&GAN   & 77.12      \\ 
		-w/o both above modules & 75.40 \\\hline
	\end{tabular}
	\caption{Ablation study on develop set.}
	\label{tab:ablation-study}
\end{table}

\begin{figure}[t]
	\includegraphics[width=0.48\textwidth]{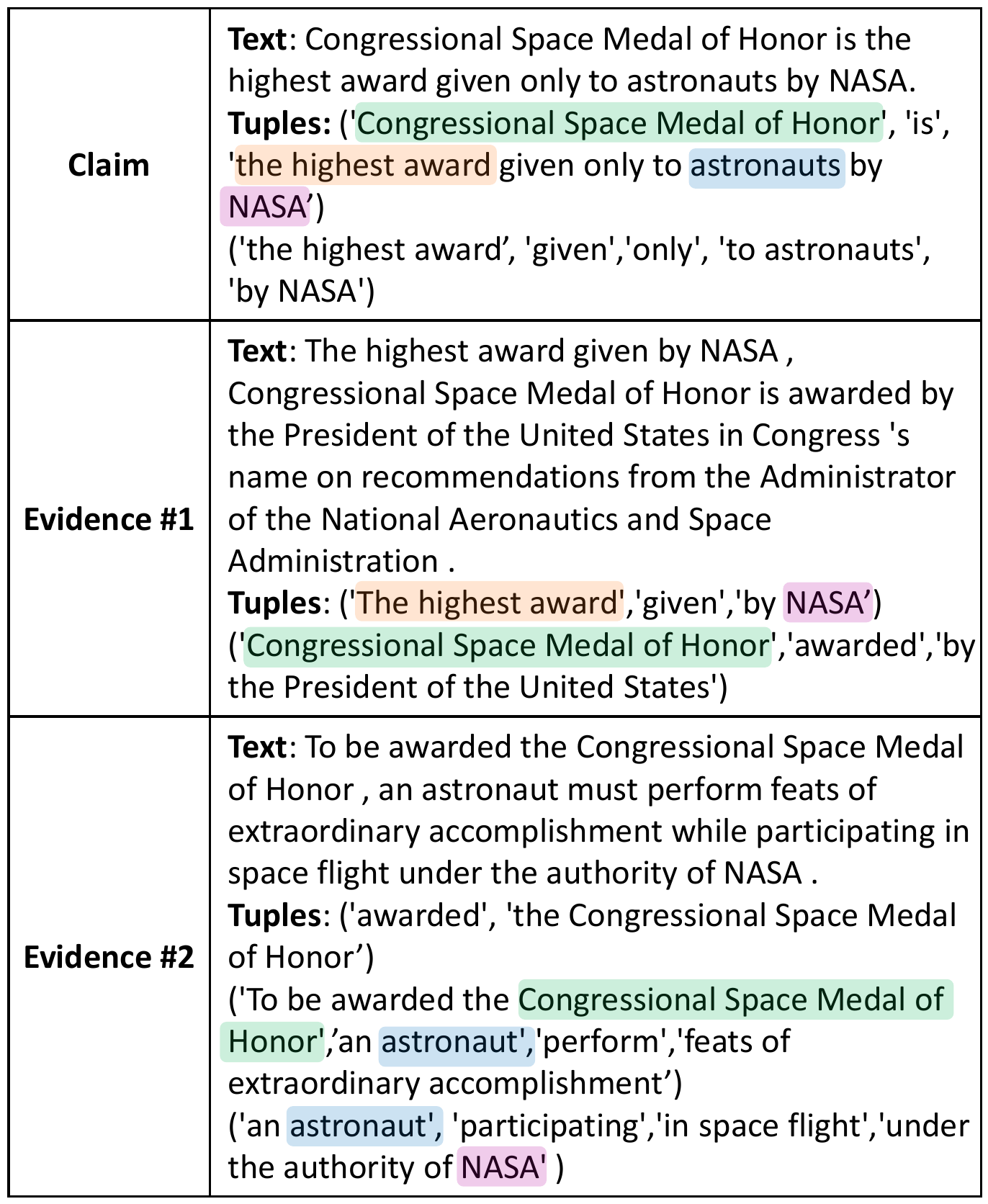}
	\caption{A case study of our approach. Facts shared across the claim and the evidence are highlighted with different colors.}
	\label{fig:model-output}
\end{figure}

As shown in Table \ref{tab:ablation-study}, compared to the XLNet baseline, incorporating both graph-based modules brings 3.76\% improvement on label accuracy.
Removing the graph-based distance drops 0.81\% in terms of label accuracy.
The graph-based distance mechanism can shorten the distance of two closely-linked nodes and help the model to learn their dependency.
Removing the graph-based reasoning module drops 2.04\% because graph reasoning module captures the structural information and \mbox{performs} deep reasoning about that.
Figure \ref{fig:model-output} gives a case study of our approach.


\subsection{Error Analysis}
We randomly select 200 incorrectly predicted instances and summarize the primary types of errors.
\par
The first type of errors is caused by failing to match the semantic meaning between phrases that describe the same event. For example, the claim states \textit{``Winter's Tale is a book"}, while the evidence states \textit{``Winter 's Tale is a 1983 novel by Mark Helprin"}. The model fails to realize that ``novel" belongs to ``book" and states that the claim is refuted. Solving this type of errors needs to involve external knowledge (e.g. ConceptNet \cite{speer2017conceptnet}) that can indicate logical relationships between different events.
\par
The misleading information in the retrieved evidence causes the second type of errors. 
For example, the claim states ``\textit{The Gifted is a movie}", and the ground-truth evidence states ``\textit{The Gifted is an upcoming American television series}". However, the retrieved evidence also contains ``\textit{The Gifted is a 2014 Filipino dark comedy-drama movie}", which misleads the model to make the wrong judgment.

\section{Related Work}

In general, fact checking involves assessing the truthfulness of a claim.
In literature, a claim can be a text or a 
subject-predicate-object triple \cite{nakashole2014language}.
In this work, we only consider textual claims.
Existing datasets differ from data source and the type of supporting evidence for verifying the claim.
An early work by \newcite{vlachos2014fact} constructs 221 labeled claims in the political domain from POLITIFACT.COM and CHANNEL4.COM, giving meta-data of the speaker as the evidence.
POLIFACT is further investigated by following works, 
including \newcite{ferreira2016emergent} who build Emergent with 300 labeled rumors and about 2.6K news articles,
\newcite{wang2017liar} who builds LIAR with 12.8K annotated short statements and six fine-grained labels, 
and 
\newcite{rashkin2017truth} who collect claims without meta-data while providing 74K news articles.
We study FEVER \cite{thorne2018fever}, which requires aggregating information from multiple pieces of evidence from Wikipedia for making the conclusion. 
FEVER contains 185,445 annotated instances, which to the best of our knowledge is the largest benchmark dataset in this area. 

The majority of participating teams in the FEVER challenge \cite{thorne2018fact} use the same pipeline consisting of three components, namely document selection, evidence sentence selection, and claim verification.
In document selection phase, participants typically extract named entities from a claim as the query and use Wikipedia search API.
In the evidence selection phase, participants measure the similarity between the claim and an evidence sentence candidate by training a classification model like Enhanced LSTM \cite{chen2016enhanced} in a supervised setting or using string similarity function like TFIDF without trainable parameters. \citet{padia-etal-2018-team} utilizes semantic frames for evidence selection.
In this work, our focus is the claim classification phase. 
Top-ranked three systems aggregate pieces of evidence through concatenating evidence sentences into a single string \cite{nie2019combining}, classifying each evidence-claim pair separately, merging the results \cite{yoneda2018ucl}, and encoding each evidence-claim pair followed by pooling operation \cite{hanselowski2018ukp}.
\newcite{zhou-etal-2019-gear} are the first to use BERT to calculate claim-specific evidence sentence representations, and then develop a graph network to aggregate the information on top of BERT, regarding each evidence as a node in the graph. 
Our work differs from \newcite{zhou-etal-2019-gear} in that (1) the construction of our graph requires understanding the syntax of each sentence, which could be viewed as a more fine-grained graph, and (2) both the contextual representation learning module and the reasoning module have model innovations of taking the graph information into consideration. 
Instead of training each component separately, \newcite{yin2018twowingos} show that joint learning could improve both claim verification
and evidence selection.

\section{Conclusion}
In this work, we present a graph-based approach for fact checking.
When assessing the veracity of a claim giving multiple evidence sentences, 
 our approach is built upon an automatically constructed graph, which is derived based on semantic role labeling. 
To better exploit the graph information, we propose two graph-based modules, one for calculating contextual word embeddings using graph-based distance in XLNet, and the other for learning representations of graph components and reasoning over the graph.
Experiments show that both graph-based modules bring improvements and our final system is the state-of-the-art on the public leaderboard by the time our paper is submitted.

Evidence selection is an important component of fact checking as finding irrelevant evidence may lead to different predictions. A potential solution is to jointly learn evidence selection and claim verification model, which we leave as a future work.
\section*{Acknowledge}
Wanjun Zhong, Zenan Xu, Jiahai Wang and Jian Yin are supported by the National Natural Science Foundation of China (U1711262, U1611264,U1711261,U1811261,U1811264, U1911203), National Key R\&D Program of China (2018YFB1004404), Guangdong Basic and Applied Basic Research Foundation (2019B1515130001),  Key R\&D Program of Guangdong Province (2018B010107005).  The corresponding author is Jian Yin.
\bibliography{Bibliography-File}
\bibliographystyle{acl_natbib}
\appendix
\section{Typology Sort Algorithm}
\label{sort-algirithm}
\begin{algorithm}[h]
	\centering
	\footnotesize
	\begin{algorithmic}[1]
		\Require
		A sequence of nodes $S = \{s_i, s_2, \cdots, s_n\}$; A set of relations  $R = \{r_1, r_2, \cdots, r_m\}$
		\Function{dfs}{node, visited, sorted\_sequence} 
		\For{each child $s_c$ in node's children}
		\If{$s_c$ has no incident edges and visited[$s_c$]==0}
		\State
		visited[$s_c$]=1
		\State
		DFS($s_c$, visited)
		\EndIf
		\EndFor
		\State
		sorted\_sequence.append(0, node)
		\EndFunction
		\State
		sorted\_sequence = []
		\State
		visited = [0 for i in range(n)]
		\State
		S,R = changed\_to\_acyclic\_graph(S,R)
		\For{each node $s_i$ in $S$}
		\If{$s_i$ has no incident edges and visited[i] == 0}
		\State
		visited[i] = 1
		\For{each child $s_c$ in $s_i$'s children}
		\State
		DFS($s_c$, visited, sorted\_sequence)
		\EndFor
		\State
		sorted\_sequence.append(0,$s_i$)
		\EndIf
		\EndFor
		\State
		\Return sorted\_sequence
	\end{algorithmic}
	\caption{Graph-based Distance Calculation Algorithm.}
	\label{alg:training}
\end{algorithm}

\section{FEVER}
\label{appendix:data-statistic}
The statistic of FEVER is shown in Table \ref{table:dataset-detail}. 
\begin{table}[h]\small
	\centering
	\begin{tabular}{c|ccc}
		\hline
		\textbf{Split} & SUPPORTED & REFUTED & NEI    \\ \hline
		Training       & 80,035    & 29,775  & 35,659 \\
		Dev            & 6,666     & 6,666   & 6,666  \\
		Test           & 6,666     & 6,666   & 6,666  \\ \hline
	\end{tabular}
	\caption{Split size of SUPPORTED, REFUTED and NOT ENOUGH INFO (NEI) classes in FEVER.}
	\label{table:dataset-detail}
\end{table}

FEVER score is calculated with equation \ref{equ:fever-score}, where 
$y$ is the ground truth label, $\hat{y}$ is the predicted label, 
$\bm{E}=[E_1,\cdots,E_k]$ is a set of ground-truth evidence,  and $\bm{\hat{E}}=[\hat{E}_1,\cdots,\hat{E}_5]$ is a set of predicted evidence.
\begin{equation}
\begin{aligned}
Instance\_Correct(y,\hat{y},\bm{E},\bm{\hat{E}}) \overset{def} {=} &\\y= \hat{y}\wedge(y=NEI\vee Evidence\_Correct(\bm{E},\bm{\hat{E})})
\end{aligned}
\label{equ:fever-score}
\end{equation}

\section{Evidence Selection Results}
\label{appendix:evidence-selection}
In this part, we present the performance of the sentence-level evidence selection module that we develop with different backbone. 
We take the concatenation of claim and each evidence as input, and take the last hidden vector to calculate the score for evidence ranking. 
In our experiments, we try both RoBERTa and XLNet. 
From Table \ref{tab:evidence-table}, we can see that RoBERTa performs slightly better than XLNet here. When we submit our system on the leaderboard, we use RoBERTa as the evidence selection model.
\begin{table}[h]
	\small
	\begin{tabular}{l|ccc|ccc}
		\hline
		\multirow{2}{*}{Model} & \multicolumn{3}{c|}{Dev. Set} & \multicolumn{3}{c}{Test Set} \\ \cline{2-7} 
		& Acc.     & Rec.    & F1    & Acc.     & Rec.    & F1    \\ \hline
		XLNet       &   26.60       &   87.33  & 40.79   &  25.55 & 85.34 & 39.33 \\
		RoBERTa      & 26.67 & 87.64 &  40.90 & 25.63 & 85.57 & 39.45 \\ \hline
	\end{tabular}
	\caption{Results of evidence selection models.}
	\label{tab:evidence-table}
\end{table}

\section{Training Details}
In this part, we describe the
training details of our experiments. We employ cross-entropy loss as the loss function. We apply AdamW as the optimizer for model training. 
For evidence selection model, we set learning rate as 1e-5, batch size as 8 and maximum sequence length as 128.

In claim verification model, the XLNet network and graph-based reasoning network are trained separately. We first train XLNet and then freeze the parameters of XLNet and train the graph-based reasoning network. We set learning rate as 2e-6, batch size as 6 and set maximum sequence length as 256. We set the dimension of node representation as 100.

\end{document}